# A Knowledge-poor Pronoun Resolution System for Turkish

**Dilek Küçük and Meltem Turhan Yöndem**

Department of Computer Engineering
Middle East Technical University
Ankara, Turkey
{dkucuk,mturhan}@ceng.metu.edu.tr

**Abstract**

A pronoun resolution system which requires limited syntactic knowledge to identify the antecedents of personal and reflexive pronouns in Turkish is presented. As in its counterparts for languages like English, Spanish and French, the core of the system is the constraints and preferences determined empirically. In the evaluation phase, it performed considerably better than the baseline algorithm used for comparison. The system is significant for its being the first fully specified knowledge-poor computational framework for pronoun resolution in Turkish where Turkish possesses different structural properties from the languages for which knowledge-poor systems had been developed.

## 1. Introduction

Anaphora resolution (AR) has various aspects including syntax, semantics and discourse. The latter two aspects require utilization of huge amounts of semantic and discourse information which is infeasible due to the inefficiency of enumerating them. As stated in (Mitkov, 2002), even most of the existing syntactic tools are not mature enough to be directly integrated into AR systems, thereby leading researchers in the area to the development of practical systems which use limited syntactic, semantic and domain knowledge. Examples of this "knowledge-poor" approach include the studies of Kennedy and Boguraev (1996), Baldwin (1996), and Mitkov (1998) for English. The approach also proved to be useful for other languages as the studies of Palomar et al. (2001) for Spanish, Trouilleux (2002) for French, and Tanev and Mitkov (2002) for Bulgarian demonstrated.

Turkish is a non-configurational language as opposed to most of the languages for which knowledge-poor anaphora resolution systems had been developed. Pro-drop is a common phenomenon in the language (as our analysis also revealed in section 2) and, Natural Language Processing (NLP) tools for Turkish, such as parsers, part-of-speech taggers, which could be integrated into other NLP systems, are not available. Therefore, it is not yet appropriate to apply a knowledge-based strategy to anaphora resolution in Turkish.

In this paper, we present a knowledge-poor pronoun resolution system for Turkish which resolves third person personal and reflexive pronouns referring to proper person names. Although there exist a number of studies on anaphora and anaphora resolution in Turkish including research on overt and zero representations of anaphora in Turkish (Enç, 1986; Erguvanlı-Taylan, 1986), a situation semantics approach to pronominal anaphora in Turkish (Tın and Akman, 1994), discourse anaphora in Turkish from the perspective of Centering Theory (Turan, 1995), resolution of dropped pronouns in Turkish (Turhan-Yöndem and Şehitoğlu, 1997), anaphora generation in Turkish (Yüksel and Bozşahin, 2002) and a computational model for pronoun resolution (Tüfekçi and Kılıçarslan, 2005) which uses Hobbs' naive approach (Hobbs, 1978), to our knowledge, our system is the first fully specified knowledge-poor computational framework for pronoun resolution in Turkish.

The paper is organized as follows: in Section 2, we present an overview of the pronoun resolution system, where we particularly describe the sample text analysis we performed before system implementation, then the way the system operates and we complete this section with the descriptions of the constraints and preferences for Turkish. We present and discuss the evaluation results of running the system against a baseline algorithm in Section 3. Conclusions and future work are presented in Section 4.

## 2. System Overview

The scope of the system is third person personal pronouns, '*o*' (he/she) and '*onlar*' (they), and reflexive pronouns, '*kendi*' (himself/herself), '*kendisi*' (himself/herself) and '*kendileri*' (themselves) and inflections of these pronouns.

Prior to the implementation of the system, a Turkish child narrative (Ilgaz, 2003a) including 8641 words with 455 third person personal and reflexive pronouns was analyzed where 285 of these pronouns were referring to proper person names.

The analysis revealed that 7% of all personal and reflexive pronouns in the examined text refer to antecedents located in the same sentence as the pronouns. 61% of the remaining pronouns corefer with antecedents residing in the previous sentence, whereas 9% of the pronouns have their antecedents in two sentences back and 4% of them have antecedents located in three sentences back. We have determined our search scope for candidates as the sentence containing the pronoun and three preceding sentences, owing to the fact that 81% of all the personal and reflexive pronouns refer to antecedents in this part of the text.

Language-specific constraints (section 2.1) and preferences (section 2.2) for Turkish were explored and proposed for the first time in the course of this analysis. As in existing systems, constraints serve to eliminate inappropriate candidates while preferences are used to sort the remaining ones.

Findings of the analysis were verified by conducting a questionnaire of 17 questions on 48 native speakers from different age, gender and job groups, and none of them were linguists. Statistical analysis of questionnaire results was carried out using Cochran's Q statistics (Cochran 1950).

After this statistical analysis, preference scores were determined by training an artificial neural network (a perceptron) where at the beginning of this learning procedure, each preference is assigned a non-optimized score of +1.

Training of the perceptron was performed using delta rule. It was trained on a test corpus including sentences taken from the questionnaire that was applied to native Turkish speakers as well as from the sample Turkish text that was analyzed. Final score for each preference is presented in parentheses in section 2.2, these optimized scores are also summarized in Table 2.1.

| Preference | Score |
|---|---|
| Quoted/Unquoted Text Preference | +2.20 |
| Recency Preference | +2.15 |
| Nominative Case Preference | +1.85 |
| First NP Preference | +1.40 |
| Nominal Predicate Preference | +1.20 |
| Repetition Preference | +1.20 |
| Punctuation Preference | +1.15 |
| Antecedent of Zero Pronoun Preference | +1.05 |

Table 2.1 Optimized Preference Scores

Input texts to the system are preprocessed to annotate the overt and zero pronouns, manually. This procedure is employed for two reasons: Our sample text analysis revealed that 74.7% of all pronouns in the text are zero pronouns, that is, pro-drop is a common phenomenon in the language. A complete pronoun resolution system should resolve these pronouns as well as the overt ones however; a parser which could identify these zero pronouns is not currently available. Secondly, automatically extracted overt pronouns may not only be non-anaphoric[1] but also be referring to an entity other than a proper person name in the text and resolution of such pronouns is out of the scope of our system. Moreover, extracted pronouns can be part of a lexical noun phrase anaphor[2] but such an extraction would not be a complete extraction and even if it had been, such anaphora is again beyond the scope of our system.

The system resolves considered pronouns by employing the strategy used in Mitkov (1998). This strategy, also presented in Mitkov (2002) as the main stages of automatic resolution of anaphors, consists of the following steps:

1. Identification of the anaphors to be resolved.

2. Location of the candidates for antecedents.

3. Selection of the antecedent on the basis of language-specific constraints and preferences.

Following these steps, the system operates as follows: At the pronoun extraction stage, simply, pronouns that are marked during preprocessing are extracted from the input text. For each extracted pronoun, proper person names in the search scope are used to form the candidate list for the antecedent of the pronoun. Particularly, in the sentence containing the pronoun, proper person names to the left of the pronoun are extracted as candidates, that is, the system does not attempt to resolve *cataphora*. A dictionary of Turkish proper person names consisting of 9060 entries is incorporated into the system in order to check the validity of an extracted proper person name; however there are still problems because of those person names which may well be used as ordinary words. Similar proper name problems in English are addressed in (Mitkov, 2002). After this stage, the system applies the constraints to discard those candidates which are inappropriate. Next, preferences are applied to the candidates by assigning them the scores of the preferences they satisfy, following the relevant studies (Lappin and Leass, 1994; Kennedy and Boguraev, 1996; Mitkov, 1998). As the final step, the system proposes the candidate with the highest aggregate score as the antecedent. In case of a tie, most recent candidate is chosen as the antecedent. If there exists no candidate for a pronoun, the system reports this pronoun as *ambiguous*.

The system resolves pronouns in the input text from left to right and when a pronoun is resolved, it is replaced with its antecedent. When all considered pronouns are resolved, the system outputs a paraphrased version of the input text in which all pronouns are replaced with their proposed antecedents.

## 2.1. Constraints

As pointed out in (Mitkov, 2002), common constraints for English include number and gender agreement, c-command constraints and selectional restrictions. Constraints for Turkish, presented below, are quite similar to those for English except for gender agreement. The latter two were derived from c-command constraints

---

[1] To our knowledge, the third person singular pronoun 'o' in Turkish can be used non-anaphorically in two cases: the first case is the idiomatic phrases like *sözüm ona* (seemingly) (i)

(i) Sözüm ona, Ayşe, Ayla'yı kandıracak.
   Word-POSS she/he-DAT Ayşe Ayla-ACC deceive-FUT
   'Seemingly, Ayşe will deceive Ayla.'

The second case is the noun phrases which contain some of the inflected forms of the noun *on* (ten) such as *onu* (on-u = ten-POSS) which are homonymous to some of the inflected forms of the pronoun like *onu* (o-nu = she/he-ACC) (ii).

(ii) Kitapların onu kayıp.
    Book-PLU-GEN ten-POSS lost.
    'Ten of the books are lost'

[2] In Turkish, the bare form of third person singular pronoun is homonymous to the demonstrative 'o' (that) which could be used to form lexical noun phrase anaphora when used before a common noun as in noun phrases like 'o çocuk' (*that child*) (i):

(i) O çocuğu görmedim.
   That child-DAT see-NEG-PAST-PERS
   (I) did not see that child.

(Mitkov, 2002). Although selectional restrictions is an applicable constraint, it was not implemented since it requires considerable semantic knowledge.

1. *Number Agreement*
   This constraint requires that a pronoun and its antecedent must agree in number.

   *Ayşe$_i$ okula gitti.*
   Ayşe school-DAT³ go-PAST

   *[Ahmet ve Fatma]$_j$ onu$_i$ gördü.*
   [Ahmet and Fatma] she-ACC see-PAST

   *Ø$_j$ Ona$_i$ el salladılar.*
   She-DAT hand wave-PAST-PERS

   'Ayşe$_i$ went to school. [Ahmet and Fatma]$_j$ saw her$_i$. (They)$_j$ waved hand to her$_i$.'

   Concerning plural pronouns, the system expects candidates for these pronouns to be in plural form or consist of multiple proper person names joined with 've' ('and') or 'ile' ('with'). When candidates in above forms are unavailable, set generation is used (Rich and Luperfoy, 1988). Proper person names in the same sentence are joined with 've' ('and') forming a set which is extracted as a candidate for a plural pronoun.

2. *Reflexive Pronoun Constraint*
   Reflexive pronoun constraint requires that the antecedent of a reflexive pronoun is the closest candidate to the pronoun.

   *Ali$_i$ kendine$_i$ güvenir.*
   Ali himself-DAT trust-AOR
   'Ali$_i$ trusts himself$_i$.'

3. *Personal Pronoun Constraint*
   This constraint requires that in a simplex sentence, a personal pronoun cannot coexist with its antecedent.

   *Ayşe$_i$ onu$_j$ gördü.*
   Ayşe she-ACC see-PAST
   'Ayşe$_i$ saw her$_j$.'

   In the sample text analysis, it was observed that only 2.4% of all personal pronouns refer to antecedents that are located in the same sentence as the pronouns. Since this percentage is negligible, this constraint is extended so that it is applied to all personal pronouns in all types of sentences.

## 2.2. Preferences

Preferences applicable to Turkish are provided below. First noun phrase preference had already been used for Bulgarian (Tanev and Mitkov, 2002). Predicate nominal preference is similar to the 'existential emphasis' preference used for English (Kennedy and Boguraev, 1996). Similarly, recency, repetition, syntactic parallelism and subject preferences were extensively used for different languages (Mitkov, 1998; Trouilleux, 2002). To our knowledge, three of them, namely, *quoted/unquoted text, punctuation,* and *antecedent of zero pronoun* preferences were not used in any other system before.

Although they are also applicable, *subject* and *syntactic parallelism* preferences were not implemented since they require considerable syntactic knowledge. However, in order to make partial use of subject preference at the same time keeping system's knowledge-poor nature, a special case of this preference, namely *nominative case* preference was implemented in the system.

1. *Quoted/Unquoted Text Preference*
   If a pronoun is in quoted text, it is very likely that its antecedent is also in quoted text. Similarly, if a pronoun is in unquoted text, it is very likely that its antecedent is also in unquoted text (score: +2.20).

   *"Bugün Ayşe$_i$'yi gördüm"*
   Today Ayşe-ACC see-PAST

   *dedi Zerrin.*
   say-PAST Zerrin.

   *"Ben de onu$_i$ dün*
   I too she-ACC yesteday

   *görmüştüm" dedi Murat.*
   see-PAST-PAST say-PAST Murat.

   '"(I) saw Ayşe$_i$ today." said Zerrin. "I had seen her$_i$ yesterday too" said Murat.'

2. *Recency Preference*
   This preference given to candidates in closer sentences to the sentence containing the pronoun (score: +2.15).

   *Ali oyun oynuyordu.*
   Ali game play-PROG-PAST

   *Murat$_i$ da geldi.*
   Murat too come-PAST

   *Ø$_i$ Oyunu sevdi.*
   Game-ACC like-PAST

   'Ali was playing a game. Murat$_i$ came too. (He)$_i$ liked the game.'

---

³ The following abbreviations are used throughout the paper: ACC: Accusative, AOR: Aorist, DAT: Dative, FUT: Future, GEN: Genitive, LOC: Locative, PAST: Past, PERS: Person, POSS: Possessive, PROG: Progressive

3. *Nominative Case Preference*
   This preference is given to the candidates which are in nominative case (score: +1.85).

   *"Günaydın"   dedi    $Murat_i$.*
   "Good morning" say-PAST Murat.

   *Ali  $ona_i$   baktı.*
   Ali  he-DAT look-PAST.

   '"Good Morning" said $Murat_i$. Ali looked at $him_i$.'

4. *First Noun Phrase Preference*
   First noun phrase preference is given to a candidate if it is a sentence-initial phrase (score: +1.40).

   *$Ahmet_i$ Ali'yi   gördü.    $Ø_i$ Koştu.*
   Ahmet Ali-ACC see-PAST.   Run-PAST.
   '$Ahmet_i$ saw Ali. $(He)_i$ ran.'

5. *Predicate Nominal Preference*
   This preference is given to a candidate if it is a predicate nominal (score: +1.20).

   *Bu  çocuk  $Ali_i$'ydi.*
   This child  Ali-PAST.

   *$Ø_i$ Sinirli görünüyordu.*
      Angry  seem-PROG-PAST

   'This child was $Ali_i$. $(He)_i$ seemed angry.'

6. *Repetition Preference*
   Repetition preference is given to the candidates that are repeated in the search scope more than once (score: +1.20).

   *$Ayşe_i$ parka   gitti.*
   Ayşe park-DAT go-PAST.

   *$Ø_i$ Zeynep'le   oyun   oynadı.*
      Zeynep-WITH game  play-PAST.

   *$Ø_i$ Şarkı  söyledi.*
      Song sing-PAST

   '$Ayşe_i$ went to the park. $(She)_i$ played game with Zeynep. $(She)_i$ sang a song.'

7. *Punctuation Preference*
   This preference is given to the candidates which have a comma following them (score: +1.15).

   *Yolda    $Tekin_i$, Ali'ye    seslendi.*
   Way-LOC Tekin  Ali-DAT  call-PAST

   *$Ø_i$ Çok  yorgundu.*
      Very tired-PAST.

   'On the way $Tekin_i$ called Ali. $(He)_i$ was very tired.'

   The effect of punctuation mark usage to resolution of pronominal anaphora is explained in (Say and Akman, 1996).

8. *Antecedent of Zero Pronoun Preference*
   If a zero pronoun is considered, this preference is given to the candidates that were antecedents of zero pronouns in previous sentences (score: +1.05).

   *$Ø_i$ Eve      yürüdü.*
      Home-DAT walk-PAST.

   *$Ø_i$ Kapıda    durdu.*
      Door-LOC stop-PAST.

   *$Ø_i$ Kapıyı    çaldı.*
      Door-ACC knock-PAST

   '$(He)_i$ walked home. $(He)_i$ stopped at the door. $(He)_i$ knocked the door.'

## 3. Evaluation

The system was evaluated on two different samples against a baseline algorithm favoring the most recent candidate after the application of the constraints. In the first experiment, a sample text from Metu Turkish Corpus (Say et al., 2002) and in the second experiment a Turkish child narrative (Ilgaz, 2003b) were used. These samples were selected since they contain considerable number of pronouns referring to proper person names therefore leading to more reasonable results.

With an intention to test the system on a syntactically annotated text, samples from Metu-Sabancı Turkish Treebank (Oflazer et al., 2003) were examined, but unfortunately these samples could not be used for this purpose since they do not contain sufficient number of pronouns that our system considers. Moreover, since we know of no other knowledge-poor anaphora resolution systems for Turkish, we cannot compare the evaluation results of our system with other systems.

### 3.1. Experiment 1

The sample text used in this experiment is taken from Metu Turkish Corpus (Say et al., 2002) contains 4140 words with 190 marked pronouns after preprocessing. 89.5% (170/190) of these pronouns were personal and 10.5% (20/190) of them were reflexive pronouns. 35.3% (67/190) of the pronouns were overt and 64.7% (123/190) of them were zero pronouns.

The results of running the system and the baseline algorithm on this text are presented in Table 3.1 where recall and precision are calculated using the following formulae:

*Recall = Number of pronouns correctly resolved /
   Number of pronouns identified*

*Precision = Number of pronouns correctly resolved /
   Number of pronouns attempted*

.

|           | Baseline Algorithm | Knowledge-poor System |
|-----------|--------------------|-----------------------|
| Recall    | 68.4%              | 85.3%                 |
| Precision | 70.6%              | 88%                   |

Table 3.1 Results of the First Experiment

The system correctly resolved 162 of 190 pronouns in this experiment. When incorrect resolutions were analyzed, it was observed that 15 of them were due to inconvenience of the personal pronoun constraint in these cases. In 6 of the remaining 13 cases, there was no candidate in the search scope and the last 7 cases were due to reasons such as extraction of non-proper names as candidates and semantic issues.

### 3.2. Experiment 2

The sample text used in this experiment is a Turkish child narrative (Ilgaz, 2003b) of 11315 words including 205 pronouns after preprocessing. 92.7% (190/205) of all pronouns were personal and 7.3% (15/205) of them were reflexive pronouns; 76.1% (156/205) of the pronouns were zero and 23.9% (49/205) of them were overt pronouns.

The results of running the system and baseline algorithms on this text is presented in Table 3.2.

|           | Baseline Algorithm | Knowledge-poor System |
|-----------|--------------------|-----------------------|
| Recall    | 65.8%              | 73.7%                 |
| Precision | 81.3%              | 91%                   |

Table 3.2 Results of the Second Experiment

In this experiment, antecedents of 151 pronouns were correctly identified by the system. 39 of 54 failures were due to non-existence of the correct antecedent in the search scope. Remaining 15 cases have reasons such as proper name extraction problems and semantic issues as in the first experiment.

As these experiments demonstrated, the performance of the knowledge-poor system is better than the baseline algorithm. Still, some improvements can be made to increase the success rate of the system by considering the sources of incorrect resolutions in the experiments.

### 4. Conclusion

In this paper, we present a knowledge-poor pronoun resolution system for Turkish which uses limited syntactic knowledge to identify the antecedents of third person personal and reflexive pronouns in Turkish.

The system is compared against a baseline algorithm favoring the most recent candidate on two different text samples. The evaluation results demonstrated that the system performs considerably better than the baseline algorithm. To our knowledge, the system is the first fully specified knowledge-poor computational framework for pronoun resolution in Turkish; thereby it provides evidence for the applicability of the knowledge-poor approach to Turkish.

As further studies, the system can be extended to resolve pronouns with noun phrase (NP) antecedents and it can be made to execute in fully automated mode by extending it with the ability to detect overt and zero pronouns that it will attempt to resolve. This latter improvement could be achieved by integrating the system with a successful parser for Turkish to detect zero pronouns.